\documentclass{article}

\usepackage{arxiv}

\usepackage[utf8]{inputenc} % allow utf-8 input
\usepackage[T1]{fontenc}    % use 8-bit T1 fonts
\usepackage{hyperref}       % hyperlinks
\usepackage{url}            % simple URL typesetting
\usepackage{booktabs}       % professional-quality tables
\usepackage{amsfonts}       % blackboard math symbols
\usepackage{nicefrac}       % compact symbols for 1/2, etc.
\usepackage{microtype}      % microtypography
\usepackage{lipsum}
\usepackage{amsmath,amssymb,amsfonts, algorithm2e}
\usepackage{algorithmic}
\usepackage{graphicx}
\usepackage{textcomp}
\usepackage{xcolor}

\title{Unsupervised Point Cloud Registration via \\
Salient Points Analysis (SPA)}
\begin{document}

\author{
  Pranav Kadam \\
  Media Communications Lab \\
  University of Southern California\\
  Los Angeles, CA, USA \\
  \texttt{pranavka@usc.edu} \\
  \And
 Min Zhang \\
  Media Communications Lab\\
  University of Southern California\\
  Los Angeles, CA, USA \\
  \texttt{zhan980@usc.edu} \\
   \And
 Shan Liu\thanks{This work was supported by Tencent Media Lab.} \\
  Tencent Media Lab\\
  Tencent America\\
  Palo Alto, CA, USA \\
  \texttt{shanl@tencent.com} \\
  \And
 C.-C. Jay Kuo \\
  Media Communications Lab\\
  University of Southern California\\
  Los Angeles, CA, USA \\
  \texttt{cckuo@sipi.usc.edu} \\
}

\maketitle

\begin{abstract}
An unsupervised point cloud registration method, called salient points
analysis (SPA), is proposed in this work. The proposed SPA method can
register two point clouds effectively using only a small subset of
salient points. It first applies the PointHop++ method to point clouds,
finds corresponding salient points in two point clouds based on the
local surface characteristics of points and performs
registration by matching the corresponding salient points. The SPA
method offers several advantages over the recent deep learning based
solutions for registration. Deep learning methods such as PointNetLK and
DCP train end-to-end networks and rely on full supervision (namely,
ground truth transformation matrix and class label). In contrast, the
SPA is completely unsupervised. Furthermore, SPA's training time and
model size are much less. The effectiveness of the SPA method is
demonstrated by experiments on seen and unseen classes and noisy point
clouds from the ModelNet-40 dataset.
\end{abstract}


\section{Introduction}\label{sec:introduction}

Point cloud registration refers to the task of finding a spatial
transformation that aligns two point cloud sets. Spatial transformation
can be rigid (e.g., rotation or translation) or non-rigid (e.g., scaling
or non-linear mappings). Reference and transformed point clouds are
called the {\em target} and the {\em source}, respectively.  The goal of
registration is to find a transformation that aligns the source with the
target. 

Point cloud registration has gained popularity with proliferation of 3D
scanning devices like LIDAR and its emerging applications in robotics,
graphics, mapping, etc. Registration is needed to merge data from
different sensors to obtain a globally consistent view, mapping a new
observation to known data, etc. Registration is challenging due to
several reasons. The source and the target point clouds may have
different sampling densities and different number of points.  Point clouds
may contain outliers and/or be corrupted by noise. Sometimes, only
partial views are available. 

%%%%%%%%%%%%%%%%%%%%%%%%%%%%%%%%%%%%%%%%%%%%%%%%%%%%%%%
\begin{figure}[htbp]
\centerline{\includegraphics[width=3in]{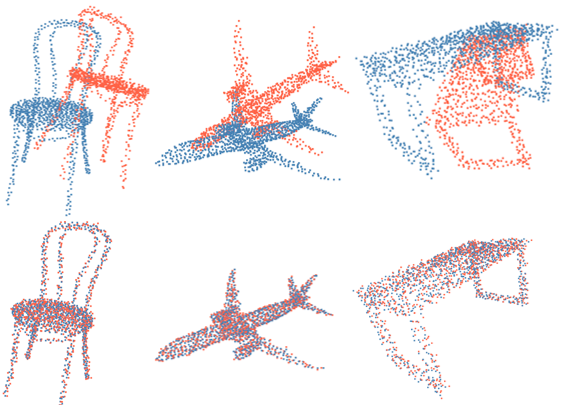}}
\caption{Registration of noisy point clouds (in orange) with noiseless
point clouds (in blue) from ModelNet-40 dataset \cite{wu20153d} using
the SPA method, where the first row is the input and the second row is 
the output.}\label{fig:examples}
\end{figure}
%%%%%%%%%%%%%%%%%%%%%%%%%%%%%%%%%%%%%%%%%%%%%%%%%%%%%%%

With the rich deep learning literature in 2D vision, a natural
inclination for 3D vision researchers is to develop deep learning
methods for point cloud processing nowadays. These solutions are heavily
supervised and trained end-to-end. Our method offers a contrasting
difference -- it does not demand ground truth transformation matrix
and/or class label for training. It is totally unsupervised.  We develop
a novel solution, called the Salient Points Analysis (SPA), that
exploits local surface properties in selecting a set of salient points.
This small subset of points are representative and can be used to
estimate the rotation and translation of the entire point cloud set
effectively and accurately (see examples in Fig.  \ref{fig:examples}).
The performance of the SPA method will be demonstrated by experiments on
seen and unseen classes and noisy point clouds from the ModelNet-40
dataset. 

The rest of this paper is organized as follows. Related work is reviewed
in Sec. \ref{sec:review}. The SPA method is described in detail in Sec.
\ref{sec:method}.  Experimental results are reported in Sec.
\ref{sec:experiments}. Finally, concluding remarks are given in Sec.
\ref{sec:conclusion}. 

\section{Related Work}\label{sec:review}

The problem of registration (or alignment) has been studied for a long
while. Prior to point cloud processing, the focus has been on aligning
lines, parametric curves and surfaces. Traditional methods rely on the
spatial coordinates of points. Recent methods adopt feature-based or
learning-based approach.

{\bf Classic Methods.} One classic algorithm for point cloud
registration is the Iterative Closest Point (ICP) method
\cite{besl1992method}. The ICP alternates between establishing point
correspondences and estimating the transformation.  For every point in
the source, a matching point in the target is found to minimize the
Euclidean distance between the two in the 3D space. Once the point
correspondences are established, the rotation matrix that minimizes the
point-wise mean square error can be computed using the singular value
decomposition (SVD). The ICP is slow and prone to local optima. It works
well for only a small degree of rotation. Several variants of ICP were
proposed later to address some of the drawbacks such as the globally
optimum ICP (Go-ICP) \cite{yang2015go} and the fast global registration
(FGR) \cite{zhou2016fast}.

%%%%%%%%%%%%%%%%%%%%%%%%%%%%%%%%%%%%%%%%%%%%%%%%%%%%%%%
\begin{figure*}[htbp]
\centerline{\includegraphics[width=7in]{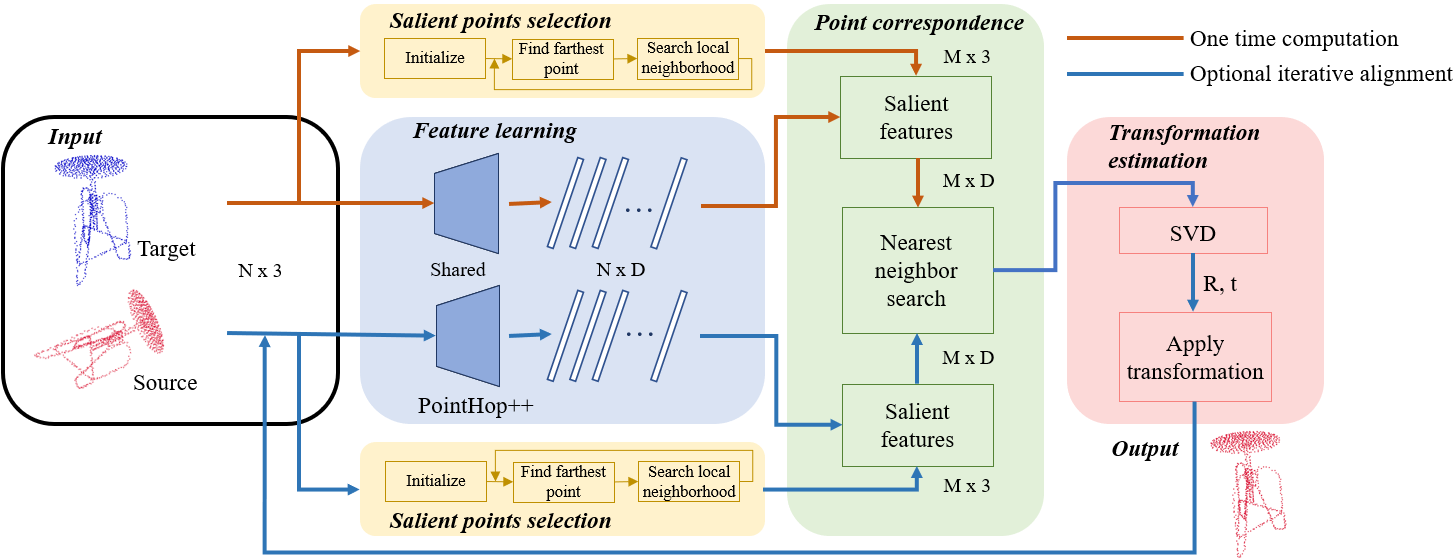}}
\caption{The system diagram of the proposed SPA method.} \label{fig:system}
\end{figure*}
%%%%%%%%%%%%%%%%%%%%%%%%%%%%%%%%%%%%%%%%%%%%%%%%%%%%%%%

{\bf Deep Learning Methods.} PointNet \cite{qi2017pointnet} was the
first deep network developed for point cloud classification and
segmentation tasks.  PointNetLK \cite{aoki2019pointnetlk} uses the
PointNet to learn the feature vector from source/target point clouds
and then finds the transformation in an iterative manner based on the
Lucas and Kanade (LK) algorithm \cite{lucas1981iterative}. The Deep
Closest Point (DCP) method \cite{wang2019deep} uses a combination of
local and global point features followed by a transformer network. The
transformer uses an attention mechanism to enable source/target point
cloud features to learn more contextual information. Then, a pointer
network is used to find point correspondences. Finally, the SVD is used
to estimate the transformation. The DCP uses point-wise features. It is
different from a feature vector for the whole point cloud adopted by
PointNetLK. Yet, both are fully supervised and trained in an end-to-end
manner. Also, they both demand a long training time and a large model
size. Our SPA method is much more favorable in all three aspects. 

{\bf PointHop and PointHop++.} An explainable point cloud classification
called PointHop was proposed in \cite{zhang2020pointhop}.  It consists
of multiple PointHop units in cascade. Each PointHop unit constructs a
vector of local attributes for every point in the point cloud. A set of
nearest neighbors of each point in the 3D space is retrieved, and the 3D
space is partitioned into eight octants centered around the point under
consideration. The centroid of input point attributes in each octant is
computed. The eight centroids are concatenated to form a new attribute
vector. The Saab transform \cite{kuo2019interpretable} is used to reduce
the dimension of the attribute vector.  Since features of different
channels at the output of the Saab transform are uncorrelated, we can
decouple them for separable transforms, leading to a more effective
transform called the channel-wise (c/w) Saab transform. The c/w Saab
transform is adopted by PointHop++ \cite{zhang2020pointhop++}, which is
an enhanced version of PointHop.  Both PointHop and PointHop++ use the
successive subspace learning (SSL) principle \cite{chen2020pixelhop} to
derive features in an unsupervised manner. PointHop++ will be adopted
in the proposed SPA method for feature learning.

\section{Proposed SPA Method}\label{sec:method}

Let $X \in {\mathbb R}^3$ and $Y \in {\mathbb R}^3$ be the target and
the source point clouds. $Y$ is obtained from $X$ through rotation and
translation, which are represented by rotation matrix, $R_{XY} \in
SO(3)$, and the translation vector, $t_{XY} \in {\mathbb R} ^3$. Given
$X$ and $Y$, the goal is to find the optimal rotation $R_{XY}^*$ and
translation $t_{XY}^*$ for $X$ so that they minimize the point-wise mean
squared error between transformed $X$ and $Y$.  The error term can be
written as
\begin{equation}\label{eq:registration}
E(R_{XY},t_{XY})=\frac{1}{N}\sum\limits_{i=0}^{N-1}||R_{XY}^*x_i
+t_{XY}^*-y_i||^2.
\end{equation}
For simplicity, $X$ and $Y$ are assumed to have the same number of
points, which is denoted by $N$. If the source and target point clouds
have a different number of points, only their common points would be
considered in computing the error. 

The system diagram of the proposed SPA method is shown in Fig.
\ref{fig:system}.  The target and the source point clouds are fed into
PointHop++ \cite{zhang2020pointhop++} to learn features. The output
for each point is a $D$-dimensional feature vector.  The salient points
selection module selects $M$ salient points from the set of $N$ points.
Features of the $M$ salient points are used to find point
correspondences using the nearest neighbor search.  Finally, SVD is used
to estimate the transformation. The output point cloud can be used as
the new source for iterative alignment. Yet, the training of PointHop++
is done only once. We will elaborate each module below. 

{\bf Feature Learning.} The target and the source point clouds are fed
into the same PointHop++ to learn features for every point.  We use
target point clouds from the training dataset to obtain Module I of
PointHop++ using a statistical approach.  The training process is same
as that in \cite{zhang2020pointhop++}. We should emphasize that Modules
II and III of PointHop++ are removed in our current system. They
include: feature aggregation and classifier. Cross-entropy-based feature selection is also
removed. These removed components are related to the classification but
not the registration task. As a result, feature learning with
PointHop++ is completely unsupervised.  Features from four Pointhop
units are concatenated to get the final feature vector.  The feature
dimension depends on energy thresholds chosen in PointHop++ design.  We
will demonstrate that our method can generalize well on unseen data and
unseen classes in Sec. \ref{sec:experiments}.

{\bf Salient Points Selection.} The Farthest Point Sampling (FPS)
technique is used in PointHop++ to downsample the point cloud. It offers
several advantages for classification such as reduced computation and
rapid coverage of the point cloud set. However, FPS is not ideal for
registration as it leads to different points being sampled from the target and
the source point clouds. We propose a novel method to select a subset of
salient points based on the local geometry information.  The new method
has a higher assurance that selected points be the same ones sampled
from the source and the target. It is detailed below.

We apply the Principal Component Analysis (PCA) to the $(x,y,z)$
coordinates of points of a local neighborhood which is centered at every
point. There are $N$ points. Thus, it defines $N$ local neighbors and we
conduct the PCA $N$ times.  Mathematically, let $p_{i,1}, p_{i,2}, ...,
p_{i,K}$ be the $K$ nearest neighbors of $p_i$. The input to this local
PCA is $K$ 3D coordinate vectors. We obtain a 3x3 covariance matrix that
has three eigenvectors and three associated non-negative eigenvalues. We
pay attention to the smallest eigenvalue among the three and denote it
with $\lambda_i$, which is the energy of the third principal component
of the local PCA at point $p_i$. 

Before proceeding, it is worthwhile to give an intuitive explanation to
this quantity.  Points lying on a flat surface can be well represented
by two coordinates in the transformed domain so that the energy of its
third component, $\lambda_i$, is zero or close to zero. For points lying
on edges between surfaces, corners or distinct local regions, they have
a large $\lambda_i$ value.  These are salient point candidates since
they are more discriminative.  Thus, we rank order all points based on
their energy values of the third principal component from the largest to
the smallest.  The point with the largest energy $\lambda$ is
initialized as the first salient point
\begin{equation}
m_0 = p_i, \mbox{  s.t. } i=\mbox{argmax}_j(\lambda_j)
\end{equation}

\begin{figure}[htbp]
\centerline{\includegraphics[width=3in]{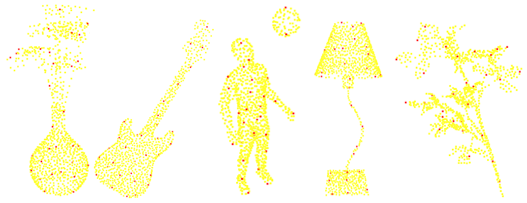}}
\caption{$M$ salient points (with $M=32$) of several point cloud sets 
selected by our algorithm are highlighted.}\label{fig:salient_point}.
\end{figure}

If we select $M$ points simply based on the rank order of $\lambda_i$,
we may end up selecting many points from the same region of largest
curvatures.  Yet, we need salient points that span the entire point
cloud. For this reason, we use the FPS to find the farthest region that
contain salient point candidates. In this new region, we again look for the point which has the largest $\lambda_i$. To be more exact, we consider two
factors jointly: 1) selecting the point according to the order of the
$\lambda_i$ value, and 2) selecting the point that has the largest
distance from the set of previously selected points.  This process is
repeated until $M$ salient points are determined. Several examples of
salient points selected by our method are shown in Fig.
\ref{fig:salient_point}. 

%%%%%%%%%%%%%%%%%%%%%%%%%%%%%%%%%%%%%%%%%%%%%%%%%%%%%%%
%%%%%%%%%%%%%%%%%%%%%%%%%%%%%%%%%%%%%%%%%%%%%%%%%%%%%%%

{\bf Point Correspondence and Transformation Estimation.} We get $M$
salient points from the target and source. Also, we obtain their features from
PointHop++ to establish point correspondences.  For each selected
salient point in the target point cloud, we use the nearest neighbor
search technique in the feature space to find its matched point in the
source point cloud. The proximity is measured in terms of the Euclidean
distance between the correspondence pair. Finally, we can form a pair
of points, $(q_m,q'_m)$, of the $m$-th matched pair. 

Once we have the ordered pairs, the next task is to estimate the
rotation matrix and translation vector that best aligns the two point
clouds. We revisit Eq. (\ref{eq:registration}), which gives the
optimization criterion.  A closed-form solution, which minimizes the
mean-squared-error (MSE), can be obtained by SVD. The step is summarized
below. First, we find centroids of two point clouds as
\begin{equation}
\Bar{x}= \frac{1}{N} \sum\limits_{i=0}^{N-1}x_i, \quad \Bar{y}=\frac{1}{N}\sum\limits_{i=0}^{N-1}y_i.
\end{equation}
Then, the covariance matrix is computed via
\begin{equation}
Cov(X,Y)=\sum\limits_{i=0}^{N-1}(x_i-\Bar{x})(y_i-\Bar{y})^T.
\end{equation}
Then, the covariance matrix can be decomposed in form of
$Cov(X,Y)=USV^T$ via SVD, where $U$ is the matrix of left singular
vectors, $S$ is the diagonal matrix containing singular values and $V$
is the matrix of right singular vectors. Then, the optimal rotation
matrix $R_{XY}^*$ and translation vector $t_{XY}^*$ are given by
\cite{schonemann1966generalized} as
\begin{equation}
R_{XY}^*=V U^T, \quad t_{XY}^*=-R_{XY}^* \Bar{x}+\Bar{y}.
\end{equation}
Then, we can align the source point cloud to the target using this
transformation. Optionally, we can use an iterative alignment method
where the aligned point cloud at the $i^{th}$ iteration is used as the
new $source$ for the $(i+1)^{th}$ iteration. Even with the iteration,
PointHop++ is only trained once. 

\section{Experiments}\label{sec:experiments}

We use the ModelNet-40 \cite{wu20153d} dataset in all experiments. It
has 40 object classes, 9,843 training samples and 2,468 test samples.
Each point cloud sample consists of 2048 points. Only 1024 points are
used in the training. Evaluation metrics include: the Mean Squared Error
(MSE), the Root Mean Squared Error (RMSE) and the Mean Absolute Error
(MAE) for rotation angles against the X, Y and Z axes in degrees and the
translation vector. To train PointHop++, we set the numbers of nearest
neighbors to 32, 8, 8, 8 in Hops \#1, \#2, \#3 and \#4, respectively.
The energy threshold of PointHop++ is set to 0.0001. The following three
sets of experiments are conducted. 

\begin{figure}[htbp]
\centerline{\includegraphics[width=3.5in]{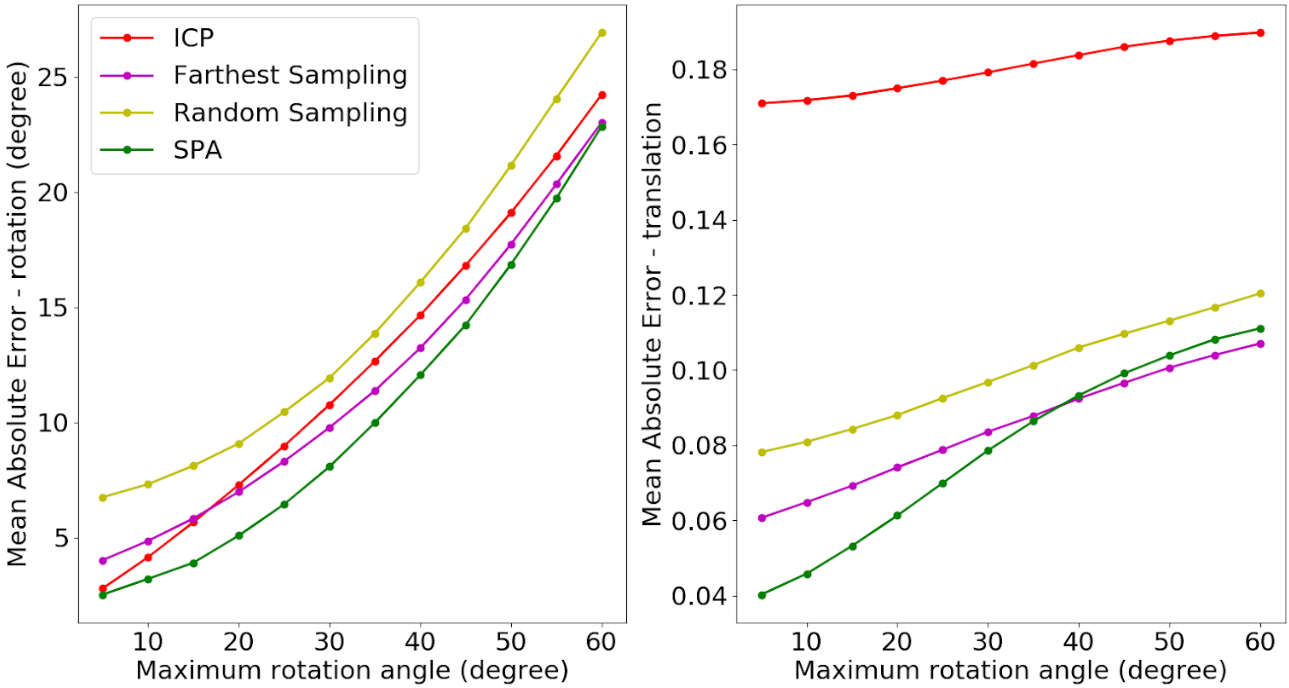}}
\caption{Comparison of the mean absolute registration errors of
rotation and translation for four benchmarking methods as a function of
the maximum rotation angle on unseen point clouds.}\label{fig:graph}
\end{figure}

{\bf Experiment \#1: Unseen point clouds.} We train PointHop++ on the target point
clouds with training samples from all 40 classes for feature learning.
Then, we apply the SPA method to test samples for the registration task.
In the experiment, we vary the maximum rotation angles against three axes from $5^{\circ}$ to $60^{\circ}$ in steps of $5^{\circ}$. For each case, the three angles are uniformly sampled between $0^{\circ}$ and the maximum angle. Translation along three axes is uniformly distributed in
$[-0.5,0.5]$. We select 32 salient points for SPA.  We replace the
salient points selection method in SPA with {\em random} selection and
{\em farthest points} selection while keeping the rest of the procedure
the same for performance comparison. We also compare the SPA method with
ICP.  For fair comparison, every point cloud undergoes exactly the same
transformation for all four benchmarking methods. The mean absolute
registration errors are plotted as a function of the maximum rotation
angle in Fig.  \ref{fig:graph}, where the results after the first
iteration are shown.  We see from Fig.  \ref{fig:graph} that the SPA
method outperforms the other three methods by a clear margin. Random
selection gives the worst results.  Interestingly, SPA with farthest
sampled points works better than ICP for larger angles.  The translation
error of all three SPA variants is much less than that of ICP.  This
validates the effectiveness of the proposed SPA method in global
registration. 

%%%%%%%%%%%%%%%%%%%%%%%%%%%%%%%%%%%%%%%%%%%%%%%%%%%%%%%%%%%%%%%%%%%%
%%%%%%%%%%%%%%%%%%%%%%%%%%%%%%%%%%%%%%%%%%%%%%%%%%%%%%%%%%%%%%%%%%%%

{\bf Experiment \#2: Unseen classes.} This experiment is used to study
the generalizability of SPA.  We adopt the same setup as that in
\cite{wang2019deep} here.  We use target point clouds from the first 20
classes of the ModelNet-40 for feature learning in the training phase.
Source and target point clouds from remaining 20 classes are used for
registration in the testing phase.  Rotation angles and the translation
are uniformly sampled from $[0^{\circ}, 45^{\circ}]$ and $[-0.5,0.5]$,
respectively. We apply SPA for 10 iterations. The results are shown in
Table \ref{tab:Result}.  We see that SPA generalizes well on categories
that it did not see during training. SPA outperforms ICP by a
significant margin.  SPA and PointNetLK have comparable performance.
Although SPA is still inferior to DCP, the difference between their
mean absolute rotation errors is less than $4^{\circ}$ while the difference
between translation errors is 0.00062. 

%%%%%%%%%%%%%%%%%%%%%%%%%%%%%%%%%%%%%%%%%%%%%%%%%%%%%%%%%%%%%%%%%%%%
\begin{table}[htbp]
\centering
\newcommand{\tabincell}[2]{\begin{tabular}{@{}#1@{}}#2\end{tabular}}
\resizebox{\textwidth}{!}{\begin{tabular}{c| c c c c c c |c c c c c c c} 

\hline 
\textbf{} &\multicolumn{6}{|c|}{\textbf{Registration errors on unseen classes}} 
&\multicolumn{6}{c}{\textbf{Registration errors on noisy input point clouds}} \\ \hline
\bf Method & \tabincell{c}{\bf MSE\\(R)}  &  \tabincell{c}{\bf RMSE\\(R)}  
& \tabincell{c}{\bf MAE\\(R)} & \tabincell{c}{\bf MSE\\(t)}  &  \tabincell{c}{\bf RMSE\\(t)}  
& \tabincell{c}{\bf MAE\\(t)} & \tabincell{c}{\bf MSE\\(R)}  &  \tabincell{c}{\bf RMSE\\(R)}  
& \tabincell{c}{\bf MAE\\(R)} & \tabincell{c}{\bf MSE\\(t)}  &  \tabincell{c}{\bf RMSE\\(t)}  
& \tabincell{c}{\bf MAE\\(t)}\\ \hline 
\tabincell{c} 
ICP \cite{besl1992method}  & 467.37 & 21.62  & 17.87 & 0.049722 & 0.222831 
& 0.186243 & 558.38 & 23.63  & 19.12 & 0.058166 & 0.241178 & 0.206283\\ \hline
PointNetLK \cite{aoki2019pointnetlk}  & 306.32 & 17.50  & 5.28 & 0.000784 & 0.028007 
& 0.007203 & 256.16 & 16.00  & 4.60 & 0.000465 & 0.021558 & 0.005652 \\ \hline
DCP \cite{wang2019deep}   & 19.20 & 4.38  & 2.68 & 0.000025 & 0.004950 & 0.003597 
&  6.93 &  2.63 & 1.52 & 0.000003 & 0.001801 & 0.001697 \\ \hline
SPA  & 354.57 & 18.83 & 6.97 & 0.000026 & 0.005120 & 0.004211 & 331.73 & 18.21 
& 6.28 & 0.000462 & 0.021511 & 0.004100 \\ \hline
\end{tabular}}
\caption{Registration performance comparison on ModelNet-40 with respect to unseen classes (left) and noisy input point
clouds (right).} \label{tab:Result}
\end{table}
%%%%%%%%%%%%%%%%%%%%%%%%%%%%%%%%%%%%%%%%%%%%%%%%%%%%%%%%%%%%%%%%%%%%

{\bf Experiment \#3: Noisy point clouds.} The objective is to understand
the error-resilience property of SPA against noisy input point cloud
data. We train PointHop++ on the noiseless target point clouds from all
40 classes for feature learning. In the testing phase, we add
independent zero-mean Gaussian noise of variance $0.01$ to the three
coordinates of each point. Rotation and
translation are uniformly sampled from $[0^{\circ},45^{\circ}]$ and
$[-0.5,0.5]$, respectively.  The results are shown in Table
\ref{tab:Result}. Again, we see that SPA outperforms ICP, has comparable
performance with PointNetLK and is inferior to DCP.  Overall, SPA is
robust to additive Gaussian noise. 

\begin{figure*}
\centerline{\includegraphics[height=1.75in]{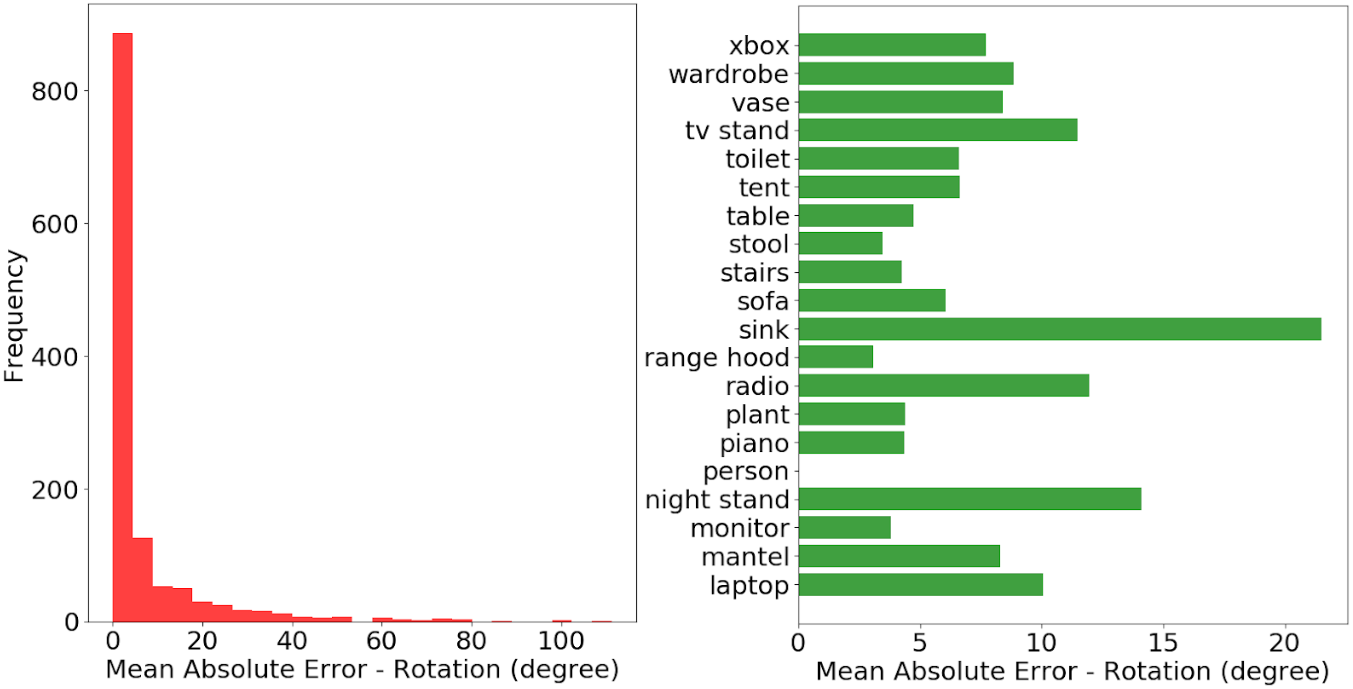}}
\caption{Histogram of mean absolute rotation error for experiment \#2 (left) and class-wise error distribution (right)}
\label{fig:histogram}
\end{figure*}

{\bf Error Analysis.} It is worthwhile to examine the error distribution
of the test data as the evaluation metrics provide only an average
measure of the performance. We plot the histogram of the MAE(R) for experiment \#2 in Fig. \ref{fig:histogram}. It is clear that a large number of point clouds align well with a very
small error. Precisely, 38\% of the test samples have their
MAE(R) less than $1^{\circ}$ and 72\% of them have their MAE(R) less
than $5^{\circ}$.

Also, in Fig. \ref{fig:histogram} we plot the class-wise MAE(R) for the 20 unseen classes that are not used during training. 8 out of 20 classes have MAE less than $5^{\circ}$. 
The error is almost $0^{\circ}$ for the $person$ class.

%%%%%%%%%%%%%%%%%%%%%%%%%%%%%%%%%%%%%%%%%%%%%%%%%%%%%%%%%%%%%%%%%%%%

%%%%%%%%%%%%%%%%%%%%%%%%%%%%%%%%%%%%%%%%%%%%%%%%%%%%%%%%%%%%%%%%%%%%

{\bf Comparison of Model Size and Training Complexity.} The model size
of SPA is only 64kB while that of DCP is 21MB.  PointNetLK trains
PointNet for the classification task first, uses its weights as the
initialization and then trains the registration
network using a new loss function. Both PointNetLK and DCP demand GPU resources to speed up
training.  However, in SPA training, CPUs are sufficient and the
training time is less than 30 minutes. Apparently, SPA offers a good
tradeoff between the model size, computation complexity and registration
performance. 

\section{Conclusion and Future Work}\label{sec:conclusion}

An unsupervised point cloud registration method called Salient Points
Analysis (SPA) was proposed in this work. The SPA method successfully
registers two point clouds with only pairs of discriminative points.
The pairing is achieved by the shortest distance in the feature domain
learned via PointHop++.  The SPA method offers comparable performance as
state-of-the-art deep learning methods at a much smaller model size with
much less training cost. It is worthwhile to conduct more thorough error
analysis for further performance improvement of the SPA method. 

\bibliographystyle{unsrt}
\bibliography{references}

\end{document}